\documentclass[journal]{IEEEtran}
\usepackage{xcolor,soul,framed} 

\colorlet{shadecolor}{yellow}
\usepackage[pdftex]{graphicx}

\graphicspath{{../pdf/}{../jpeg/}}
\DeclareGraphicsExtensions{.pdf,.jpeg,.png}

\usepackage[cmex10]{amsmath}
\usepackage{array}
\usepackage{mdwmath}
\usepackage{mdwtab}
\usepackage{eqparbox}
\usepackage{url}
\usepackage{algorithm}
\usepackage{algorithmic}
\usepackage{cite}
\usepackage{booktabs} 
\usepackage{multirow}
\usepackage{subcaption}
\usepackage{multicol}
\usepackage{amsfonts}
\usepackage{float}
\usepackage{bbm}
\usepackage{balance}
\usepackage{xcolor}
\usepackage{hyperref}

\usepackage{amssymb}
\usepackage{pifont}

\begin{document}
    \title{Do We Really Need GNNs with Explicit Structural Modeling? MLPs Suffice for Language Model Representations}
    \author{Li Zhou, Hao Jiang, Junjie Li, Zefeng Zhao, Feng Jiang, Wenyu Chen, Haizhou Li,~\IEEEmembership{Fellow,~IEEE}
    
    \thanks{Li Zhou, Hao Jiang, Zefeng Zhao, Feng Jiang, and Haizhou Li are School of Data Science, The Chinese University of Hong Kong, Shenzhen, 518172, China (email: lizhou21@cuhk.edu.cn, haojiang0729@gmail.com, zefengzhao@cuhk.edu.cn, jeffreyjiang@cuhk.edu.cn, haizhouli@cuhk.edu.cn).}
    \thanks{Junjie Li is with Department of Electrical and Eletronic Engineering, Faculty of Engineering, The Hong Kong Polytechnic University (e-mail: junjie98.li@connect.polyu.hk).}
    \thanks{Wenyu Chen is with School of Computer Science and Engineering (School of Cyber Security), University of Electronic Science and Technology of China (e-mail: cwy@uestc.edu.cn).}
}
    
\maketitle

\begin{abstract}
Explicit structural information has been proven to be encoded by Graph Neural Networks (GNNs), serving as auxiliary knowledge to enhance model capabilities and improve performance in downstream NLP tasks. 
However, recent studies indicate that GNNs fail to fully utilize structural information, whereas Multi-Layer Perceptrons (MLPs), despite lacking the message-passing mechanisms inherent to GNNs, exhibit a surprising ability in structure-aware tasks.
Motivated by these findings, this paper introduces a comprehensive probing framework from an information-theoretic perspective. 
The framework is designed to systematically assess the role of explicit structural modeling in enhancing language model (LM) representations and to investigate the potential of MLPs as efficient and scalable alternatives to GNNs.
We extend traditional probing classifiers by incorporating a control module that allows for selective use of either the full GNN model or its decoupled components—specifically, the message-passing and feature-transformation operations.
This modular approach isolates and assesses the individual contributions of these operations, avoiding confounding effects from the complete GNN architecture.
Using the Edge Probing Suite—a diagnostic tool for evaluating the linguistic knowledge encoded in LMs—we find that MLPs, when used as feature-transformation modules, consistently improve the linguistic knowledge captured in LM representations across different architectures. They effectively encode both syntactic and semantic patterns. Similarly, GNNs that incorporate feature-transformation operations show beneficial effects. In contrast, models that rely solely on message-passing operations tend to underperform, often leading to negative impacts on probing task performance.
These findings challenge conventional reliance on GNNs for structure-aware NLP tasks, underscoring the scalability and efficiency of MLP-based approaches and reinforcing the the principle of ``simpler yet stronger''.\footnote{The dataset and code are available at \href{https://github.com/lizhou21/MLPs-Beyond-GNNs}{lizhou21/MLPs-Beyond-GNNs}.}

\end{abstract}

\section{Introduction}

\IEEEPARstart{T}{he} advanced capabilities of language models (LMs)~\cite{achiam2023gpt, dubey2024llama, guo2025deepseek} in understanding and generating text have attracted significant attention across various domains, sparking widespread interest in understanding the mechanisms underlying their performance~\cite{rogers-etal-2020-primer}.
To investigate the information encoded within the internal representations of LMs, researchers have adopted various methodologies, including fill-in-the-gap probes~\cite{petroni-etal-2019-language, wu-etal-2023-plms}, analysis of self-attention weights~\cite{ravishankar-etal-2021-attention, eberle-etal-2022-transformer}, and probing classifiers that utilize LM representations as input~\cite{zhou2024mlps}. 
These studies reveal that LMs encode multiple types of knowledge, including syntactic knowledge~\cite{jawahar-etal-2019-bert, ravishankar-etal-2021-attention, cherniavskii-etal-2022-acceptability, zhang-etal-2023-well, hale-stanojevic-2024-llms}, semantic knowledge~\cite{broscheit-2019-investigating, balasubramanian-etal-2020-whats, hayashi-2024-reassessing, hayashi-2025-evaluating}, and even certain world knowledge~\cite{petroni-etal-2019-language, roberts-etal-2020-much, fierro-etal-2024-mulan, zhou2024does}.
Furthermore, certain LMs, such as BERT, have been shown to hierarchically organize knowledge within their representations, capturing surface-level details in the lower layers, syntactic structures in the middle layers, and semantic understanding in the upper layers~\cite{jawahar-etal-2019-bert, tenney-etal-2019-bert}.
Moreover, existing studies also demonstrate that explicit syntactic information contributes significantly to tasks such as machine translation~\cite{bastings-etal-2017-graph, 10.1145/3638762}, text summarization~\cite{fernandesstructured, ragazzi2024cross}, information extraction~\cite{zhou2020weighted, liu2022document, yi2025diffusyn}, and sentiment analysis~\cite{li2022dualgcn, zhang2024integrated, kanayama-etal-2024-incorporating}.



\begin{figure}[t]
    \centering
    \begin{minipage}{1\linewidth}
        \centering
        \includegraphics[width=1.0\linewidth]{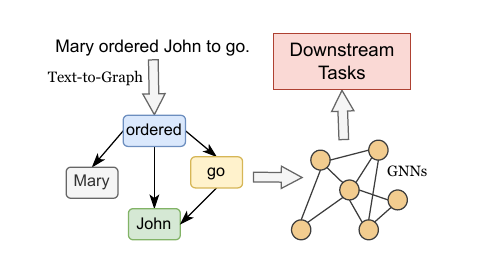}
        \subcaption{}
        \label{fig:traditional_pipeline}
    \end{minipage}

    \begin{minipage}{1\linewidth}
        \centering
        \begin{tabular}{@{}lrr@{}}
\toprule
                  & \textbf{ReTACRED} & \textbf{SemEval} \\ \midrule
\textbf{BERT}     & 87.66±0.18        & 91.07±0.26       \\
\textbf{BERT+MLPs} & 88.05±0.21        & 91.31±0.23       \\ \bottomrule
\end{tabular}
\subcaption{}
\label{tab:RE-task}
    \end{minipage}
    \caption{(a) Traditional pipeline for modeling structural information. (b) Relation extraction performance using MLPs without explicit structural information.}
    \label{fig:compare_general}
\end{figure}

To effectively model structural information in text, graph neural networks (GNNs)~\cite{Kipf:2016tc, velikovi2017graph, zhou2023dpgnn} have emerged as a powerful solution. 
Fig.~\ref{fig:traditional_pipeline} illustrates the traditional pipeline for modeling structural information, where unstructured sequential text is first parsed into structural pattern using parsing tools. These structural patterns are then processed by GNNs to capture and encode the structural information, enhancing text representations before they are fed into downstream tasks.
As the de-facto model class for representation learning on graphs, GNNs extend the architecture of multi-layer perceptrons (MLPs) with additional message-passing layers, enabling the flow of features across graph nodes. This allows GNNs to capture complex structural relationships, including long-distance dependencies in text~\cite{zhou2020weighted, zhang-etal-2018-graph, bastings-etal-2017-graph}, which are critical to understanding and representing intricate linguistic patterns.

However, recent studies highlight several limitations of GNNs, including computational inefficiency, limited scalability, and difficulties in capturing long-range dependencies~\cite{zhanggraph}. 
The effectiveness of GNNs in leveraging graph structures is also questioned, as structure-agnostic baseline models outperform GNNs on certain datasets~\cite{erricafair}, and existing knowledge-aware GNN modules often perform only simple reasoning, such as counting~\cite{wanggnn}.
These shortcomings drive interest in alternative modeling approaches, such as combining MLPs with GNNs through distillation techniques~\cite{zhang2021graph,tian2022learning, chen2024samlp}, as well as the adoption of purely graph-less methods that rely directly on MLP components~\cite{galke-scherp-2022-bag, 10.1145/3701716.3715464}.
Interestingly, research shows that even a single-layer MLP can effectively capture latent semantic information~\cite{li-etal-2023-well,zhou2024mlps}. 
As demonstrated in Fig.~\ref{tab:RE-task}~\cite{zhou2024mlps}, incorporating additional MLPs into BERT, without leveraging the explicit structural information encoded by GNNs, enhances performance on the ReTACRED~\cite{stoica2021re} and SemEval~\cite{hendrickx-etal-2010-semeval} relation extraction benchmark datasets.
Besides, MLPs have demonstrated strong performance in various tasks, including  text classification~\cite{galke-scherp-2022-bag},
citation networks~\cite{tian2022learning}, sequential recommendation~\cite{zhou2022filter, gao2024smlp4rec}, temporal set prediction~\cite{yu2023predicting}, spatio-temporal prediction~\cite{zhang2023mlpst}, and computer vision~\cite{tolstikhin2021mlp, lianmlp}.  These all-MLP architectures offer lower time complexity and fewer parameters~\cite{zhou2022filter, gao2024smlp4rec}, making them an efficient and competitive alternative across a wide range of applications.


Building on the above discussion, we propose the following research question: \textit{Do we really need GNNs with explicit structural modeling to enhance language model representations for improved performance on downstream NLP tasks?}
To address this, we develop a comprehensive probing framework grounded in an information-theoretic perspective by decoupling the micro-architecture of the GNN model. Specifically, we enhance traditional probing classifiers with a control module that selectively activates either the full GNN model or its decoupled components—namely, message-passing and feature-transformation operations. By employing this modular analysis approach, we isolate and evaluate the individual contributions of these operations, eliminating confounding effects introduced by the complete architecture.
We apply this framework to the classical Edge Probing Suite to evaluate the role of explicit structural information in improving language model representations across various types of linguistic knowledge. 
Experimental results reveal that feature-transformation operations in GNNs are the primary drivers of performance improvements, while message-passing operations contribute minimally, despite their focus on structural modeling. Furthermore, substituting GNNs with simple MLPs results in superior probing performance, indicating that explicit structural information may not be critical for enhancing language model representations in this context.

The contributions in this paper can be listed as:
\begin{enumerate}
    \item To investigate the key factors driving GNN performance, we decouple their micro-architecture into two key components: feature-transformation, modeled by an MLP module, and message-passing, captured by a dedicated Message module.
    \item Based on that, we extend traditional probing classifiers by incorporating a control module that allows for selective use of either the full GNN model or its decoupled components.
    \item Probing experiments on eight linguistic tasks from the Edge Probing Suite show that GNN performance gains primarily stem from feature-transformation operations, rather than message-passing operations that model structural information.
    \item Extensive experimental analysis demonstrates that MLPs without explicit structural modeling are sufficient for enhancing language model representations. Moreover, their recent alternative, Kolmogorov–Arnold Networks (KAN), shows comparable performance.
\end{enumerate}

\section{Preliminaries and Related works}

\subsection{Syntactic Structure Modeling}
With the increasing success of GNNs in processing structured data, significant research efforts have focused on utilizing syntactic structural information in NLP through GNNs. 
For instance, some studies~\cite{zhang-etal-2018-graph, zhou2020eawgcn} apply Graph Convolutional Networks (GCNs)~\cite{kipf2017semi} directly to pruned syntactic Universal Dependencies (UD)\cite{de-marneffe-etal-2021-universal} trees to enhance performance in relation extraction tasks. Similarly, incorporating syntactic structures into neural attention-based encoder-decoder models for machine translation has also been shown to yield effective results~\cite{bastings-etal-2017-graph}.
In the context of other generation tasks, Heterogeneous Graph Transformer (HetGT) and its variants~\cite{yao-etal-2020-heterogeneous, li-flanigan-2022-improving} have been proposed to encode Abstract Meaning Representation (AMR) graphs~\cite{banarescu-etal-2013-abstract}, achieving notable performance improvements. Furthermore, , Sachan et al.\cite{sachan-etal-2021-syntax} proposed a syntax-based graph neural network (syntax-GNN), which integrates UD structures into a modified transformer encoder by replacing the self-attention sublayer with a Graph Attention Network (GAT)~\cite{velickovic2018graph}. Their work investigates both \textit{late fusion} and \textit{joint fusion} strategies for combining syntactic information.
In addition, other studies~\cite{li2022dualgcn, zhang2024integrated} explore the use of Dual-Channel GCNs to simultaneously integrate syntactic and semantic information, particularly for sentiment analysis tasks, highlighting the potential of such approaches in capturing rich linguistic features.

However, comparisons between GNNs and structure-agnostic baseline models—such as models that do not explicitly leverage graph structural information—reveal that GNNs may fail to fully utilize graph-specific features on certain datasets~\cite{erricafair}. This challenges the widely held assumption that GNNs inherently and consistently exploit graph structural information. Furthermore, MLPs have been demonstrated as strong baseline models, capable of effectively capturing latent semantic information, even with a single-layer architecture~\cite{li-etal-2023-well, zhou2024mlps}.   
In this paper, we aim to investigate the necessity of using GNN encoders to explicitly integrate syntactic structural information in NLP tasks. Additionally, we examine the potential of employing a simplified variant of GNNs—specifically, an MLP without additional message-passing layers—as a means to enhance language model representations.

\begin{figure*}[ht]
    \centering
\includegraphics[width=1\linewidth]{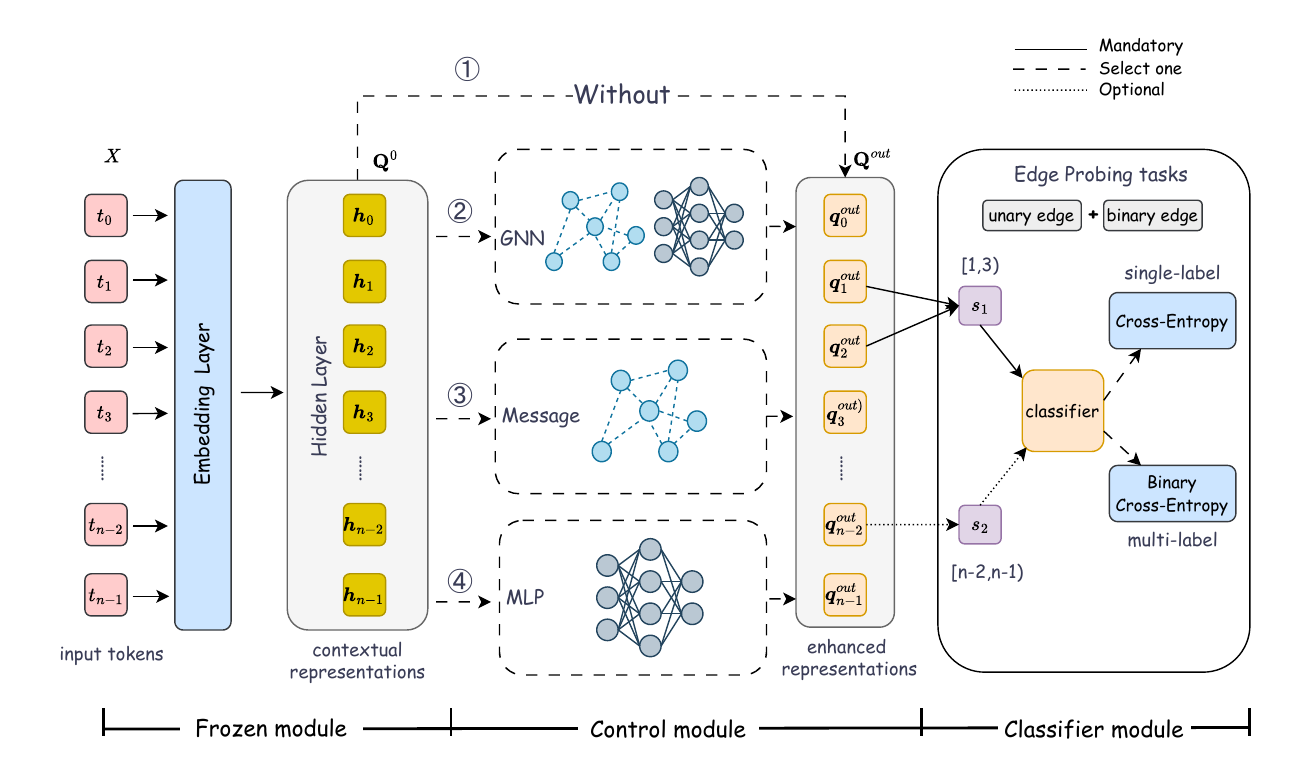}
    \caption{Probing framework. The parameters in the \textit{Frozen} module are fixed. Solid arrows denote the necessary process, while dashed arrows indicate the process configuration tailored to the specific experimental setup and dataset.}
    \label{fig:framework}
\end{figure*}

\subsection{Probing Classifiers}
Probing classifiers~\cite{belinkov-2022-probing} are a widely used method to interpret and analyze  language models in NLP. 
This approach involves training a classifier 
to predict specific linguistic properties from a model's representations, providing insights into both the models and the encoded linguistic features.

To facilitate this, we formally define the principles and process of probing classifiers. 
Let a language model be represented as $f$, which generates representations $f(X)$ encoding linguistic information, where $X$ denotes the input text provided to the model.
A probing classifier, denoted as $g:f\left( x \right) \mapsto \tilde{z}$, maps the representations
to a specific linguistic property $\tilde{z}$. 
The classifier $g$ is trained and evaluated using a probing task dataset $\mathcal{D} _{\tau}=\left\{ (x^{\left( i \right)},z^{\left( i \right)}) \right\}$,
where $i$ indexes individual examples, $x^{(i)}$ is the input text, and $z^{(i)}$ is the corresponding label.
Here, $\tilde{z}$ represents the abstract linguistic property being probed (e.g., part of speech or syntactic role), while $z^{(i)}$ denotes its specific realization for the $i$-th example.

The measured performance \textsc{Pref}$\left( g,f,\mathcal{D} _{\tau} \right)$  quantifies how well the representations in $f$ encode the linguistic property $\tilde{z}$.
From an information-theoretic perspective, the process of training the probing classifier $g$ can be understood as an estimation of the mutual information between the representations $f(X)$ and the linguistic property $\tilde{z}$~\cite{pimentel-etal-2020-information, zhu-rudzicz-2020-information}.
Previous research has shown that lower layers of language models are more effective at predicting word-level syntax, while higher layers excel at capturing sentence-level syntax and semantic knowledge~\cite{jawahar-etal-2019-bert, tenney-etal-2019-bert}. 
Furthermore, Vulic et al.\cite{vulic-etal-2020-probing} revealed that monolingually pre-trained language models encode a substantial amount of type-level lexical knowledge, which is predominantly concentrated in the lower layers of the Transformer. In addition, decoder-only LMs exhibit stronger capabilities in distinguishing machine-generated text compared to encoder-only LMs, leveraging their learned representations~\cite{cheng2024beyond}.

\section{Probing Framework}

In this paper, our focus is not limited to analyzing the inherent information encoded by LM representations. Instead, we aim to explore whether the representational capabilities of LMs can be enriched through the integration of syntactic structural information.
To achieve this, we extend the probing framework by introducing a component $c$ between $f$ and $g$ to further enhance the language model representations.
We use performance differences $\bigtriangleup$\textsc{Pref} to represent the effectiveness of component $c$ in enhancing the linguistic property $\tilde{z}$:
\begin{equation}
\bigtriangleup\textsc{\text{Pref}}=\textsc{\text{Pref}}\left( g, f, c, \mathcal{D} _{\tau} \right) -\textsc{\text{Pref}}\left(g, f, \mathcal{D} _{\tau} \right) 
\end{equation}
Based on this principle, we design a probing framework, as illustrated in Fig.~\ref{fig:framework}, which consists of three modules: the \textit{Frozen}, \textit{Control}, and \textit{Classifier} modules.
The input text is first processed by the \textit{Frozen} module, which extracts the representations learned by LMs. These representations are then refined by the \textit{Control} module to enhance their expressiveness. Finally, the enhanced representations are passed to the \textit{Classifier} module for downstream probing tasks, where performance variations are examined to assess the role of the \textit{Control} components in capturing task-specific information and their relationship to the corresponding linguistic features.


\subsection{Frozen Module}
The \textit{Frozen} module consists of a language model (LM), which is used to obtain the contextual representation of a given sentence $X= [t_{0}, t_1, \cdots, t_{n-1}]$, resulting in $f(X) = [\textbf{h}_{0}, \textbf{h}_1, \cdots, \textbf{h}_{n-1}]$, where each
$t_i$ represents a subword unit generated by the LM tokenizer, rather than linguistic tokens.
Linguistic tokens, in contrast, represent the fundamental units of language used in linguistics or natural language processing—such as words, morphemes, or phrases—that form the building blocks of language expressions. These differ conceptually and functionally from the subword units generated through LM tokenization.
In this module, unlike traditional fine-tuning approaches for LMs, all parameters remain frozen during the probing tasks. This ensures that the probing model neither modifies the internal representations of the LMs nor acquires new information from them while executing the tasks. By preserving the integrity of the knowledge and representations learned by the LMs during training, this approach ensures that any variations in performance can be attributed primarily to the components in the subsequent control module, rather than to fine-tuning the LMs.



\subsection{Control Module}

Before feeding representations generated by LMs into specific downstream probing tasks, we incorporate the \textit{Control} module as a variable component of the overall probing framework.
This module consists of three optional components: a \textit{GNN}, which is traditionally used to encode syntactic structural information, and its decoupled components—the message-passing operation, referred to as the \textit{Message} component, and the feature-transformation operation, referred to as the \textit{MLP} component.

\paragraph{GNN} 
Given a graph $G = (V, E)$, 
the operations within a typical GNN layer can be naturally associated with message-passing and feature-transformation processes:
\begin{itemize}
    \item \textbf{Message-Passing}: For each node $v_i\in V$, information is aggregated from its neighboring nodes based on the graph structure $E$. This involves applying an aggregation function to combine the features of neighboring nodes, which captures the local relational information within the graph.
    \item \textbf{Feature-Transformation}: After aggregating information in the message-passing step, node representations are updated using a learnable function, typically implemented as a Multi-Layer Perceptron (MLP) to enhance the representations and capture higher-level features.
\end{itemize}

Usually, $\mathbf{A}$ denotes the adjacency matrix of the graph, representing its topology and the connections between nodes. $\mathbf{A}_{ij}=1$ indicates an edge between node $i$ and node $j$, while $\mathbf{A}_{ij}=0$ means no edge.
In the context of text, each token is treated as a node, and the explicit syntactic structure of the sentence defines the graph connecting these nodes~\cite{zhang-etal-2018-graph, zhou2020eawgcn}.
To implement the \textit{GNN} component, we adopt the Graph Convolutional Network (GCN)~\cite{DBLP:conf/iclr/KipfW17}, which is one of the most fundamental and widely-used GNN models. 
In $l$-th layer of the GCN, node representations are updated according to the following equation:
\begin{equation}
\mathbf{Q}^{(l)}=\mathrm{ReLU} (\mathbf{\hat{D}}^{-1/2}\mathbf{\hat{A}\hat{D}}^{-1/2}\mathbf{Q}^{(l-1)}\mathbf{W}^{(l)}+\textbf{b}^{(l)})\in \mathbb{R} ^{n\times d}
\label{eq:GNN}
\end{equation}
where $\mathbf{\hat{A}}=\mathbf{A}+\mathbf{I}_n$, 
$\mathbf{I}_n$ is an $n\times n$ identity matrix used to add self-loops to the adjacency matrix, ensuring each node is connected to itself.
$\mathbf{\hat{D}}$ is the degree matrix of $\mathbf{\hat{A}}$, a diagonal matrix with 
$\hat{\mathbf{D}}_{ii}=\sum_j{\hat{\mathbf{A}}_{ij}}$.
$\textbf{Q}^{(0)}=[\textbf{h}_{0}, \textbf{h}_1, \cdots, \textbf{h}_{n-1}]$ denotes the initial node representations obtained from LMs, $\mathbf{W}^{(l)}$ and $\textbf{b}^{(l)}$ are the learnable parameters of the $l$-th GCN layer.

\paragraph{Message}
In the \textit{Message} component, only the steps of propagating and aggregating information based on the syntactic structure are preserved.
 Liu et al.~\cite{liu2020towards} decouple graph convolution operations into representation transformation and propagation, proposing the Deep Adaptive Graph Neural Network (DAGNN) to adaptively integrate information from large receptive fields. 
Inspired by this, we implement the \textit{Message} component using the propagation operation of DAGNN, defined as follows:
\begin{align}
\mathbf{Q}^{(l+1)}=\hat{\mathbf{D}}^{-1/2}\hat{\mathbf{A}}\hat{\mathbf{D}}^{-1/2}\mathbf{Q}^{(l)} &\in \mathbb{R} ^{n\times d}\label{eq:message}\\
\mathbf{Q}=\mathrm{stack}\left( \mathbf{Q}^{(0)}, \mathbf{Q}^{(1)}, \cdots, \mathbf{Q}^{(k)} \right) & \in \mathbb{R} ^{n\times \left( k+1 \right) \times d}\label{eq:stack}\\
\mathbf{S}=\sigma \left( \mathbf{Q}\boldsymbol{s} \right) & \in \mathbb{R} ^{n\times \left( k+1 \right) \times 1}\\
\tilde{\mathbf{S}}=\mathrm{reshape}\left( \mathbf{S} \right) & \in \mathbb{R} ^{n\times 1\times \left( k+1 \right)}\\
\mathbf{Q}^{out}=\mathrm{squeeze}\left( \tilde{\mathbf{S}}\mathbf{Q} \right) &\in \mathbb{R} ^{n\times d}
\end{align}
where $k$ is a hyperparameter that determines the number of hops for structural propagation, aligning with the number of layers in the \textit{GNN}, 
and $\sigma$ represents the sigmoid activation function.
$\boldsymbol{s}\in \mathbb{R}^{d\times1}$ is a trainable projection vector designed to learn the extent to which each node integrates information from different hops,
It is the only learnable parameter in the \textit{Message} component. Additionally, operations such as \textit{stack}, \textit{reshape}, and \textit{squeeze} are employed to rearrange the data dimensions to ensure compatibility during computation.

\paragraph{MLP} In the \textit{MLP} component, only feature transformation is considered, and structure information from the input sentence is not incorporated, which is define as:
\begin{equation}
\textbf{Q}^{(l+1)}=\mathrm{ReLU}(\textbf{Q}^{(l)}\textbf{W}^{(l)}+\textbf{b}^{(l)})
\label{eq:mlp}
\end{equation}

Among the three components described above, 
Eq.~\ref{eq:GNN} is formed by directly combining Eq.~\ref{eq:message} and Eq.~\ref{eq:mlp}. The \textit{Message} component defines the output of the entire component, while the \textit{GNN} and \textit{MLP} components define a single network layer operation within the component.
Following  Zhou et al.~\cite{zhou2020weighted}, residual networks \cite{he2016deep} are introduced in each layer of the \textit{GNN} and \textit{MLP} components to mitigate the vanishing gradient problem, which can align with Eq.~\ref{eq:stack}.
The final output of this control module is represented as $\mathbf{Q}^{out}=[\textbf{q}_{0}^{out}, \textbf{q}_{1}^{out}, \cdots,\textbf{q}_{n-1}^{out}]$.

\begin{table*}[ht]
\caption{Example of Edge Probing Tasks}
\resizebox{\textwidth}{!}{
\begin{tabular}{@{}lll@{}}
\toprule
\textbf{Probing Tasks}     & \textbf{Examples}                                                                  & \textbf{Labels}                          \\ \midrule
\textbf{Part-of-Speech}      & So I am [thanking]$_1$ everyone for helping me .         & VBG (Verb Gerund)                              \\
\textbf{Constituents} & I felt the interview [went well]$_1$, and started to feel confident.         & VP (Verb Phrase)                     \\
\textbf{Dependencies}  & He [makes]$_2$ some good [observations]$_1$ on a few of the pic's .                                       & obj (Direct Object)             \\
\textbf{Entities} & But even so , they have to pay [165 yuan]$_1$ in income tax.        & Money                        \\
\textbf{SRL}      & But [I]$_2$ still [needed]$_1$ to do something ; getting crazy - hysteric makes no sense. & ARG0                       \\
\textbf{Coreference}   & [The interview]$_2$ didn't last long , just two hours and [it]$_1$ was over . & True                                 \\
\textbf{SPR}      & [It]$_1$ [endorsed]$_2$ the White House strategy. . .                      & \{awareness, existed after, . . . \} \\

\textbf{Rel.(SemEval)}     & [Roundworms]$_0$ or ascarids are caused by an intestinal [parasite]$_2$ called Toxocara canis.             & Product-Producer(e1,e2)                \\ \bottomrule
\end{tabular}}
\label{tab:word_level_example}
\end{table*}
\begin{table*}[ht]
\centering
\caption{Statistics for datasets used in this paper. `u' denotes unary tasks, and `b' denotes binary tasks.}
\begin{tabular}{@{}llrrrrr@{}}
\toprule
\multirow{2}{*}{\textbf{Probing Tasks}} & \multirow{2}{*}{\textbf{Dataset Source}} & \multicolumn{3}{c}{\textbf{Split}}            & \multicolumn{1}{l}{\multirow{2}{*}{\textbf{Label}}} & \multirow{2}{*}{\textbf{Type}} \\ \cmidrule(lr){3-5}
                                        &                                          & \textbf{Train} & \textbf{Dev} & \textbf{Test} & \multicolumn{1}{l}{}                                &                                \\ \midrule
\textbf{Part-of-Speech}                 & OntoNotes-PoS                            & 1,124,410      & 138,088      & 142,877       & 48                                                  & u                              \\
\textbf{Constituents}                   & OntoNotes-Const                          & 2,144,920      & 262,367      & 270,704       & 78                                                  & u                              \\
\textbf{Dependencies}                   & English Web Treebank                     & 200,468        & 24,545       & 23,876        & 49                                                  & b                              \\
\textbf{Entities}                       & OntoNotes-NER                            & 70,495         & 9,427        & 9,301         & 18                                                  & u                              \\
\textbf{SRL}                            & OntoNotes-SRL                            & 382,212        & 46,737       & 47,397        & 138                                                 & b                              \\
\textbf{Coreference}                    & OntoNotes-Coref                          & 180,817        & 22,531       & 22,684        & 2                                                   & b                              \\
\textbf{SPR}                            & SPR-2                                    & 4,879          & 617          & 569           & 20                                                  & b                              \\
\textbf{Rel.(SemEval)}                  & SemEval 2010 Task 8                      & 8,000          & 2,717        & 2,717         & 19                                                  & b                              \\ \bottomrule
\end{tabular}
\label{tab:data_sta}
\end{table*}

\subsection{Classifier Module}

For each probing task $\tau$, a probing classifier $P_{\tau}$ is trained in the classification module. 
Specifically, we adopt the edge probing design~\cite{tenney2018what}, which focuses on token-level probing and includes a diverse set of sub-sentence tasks derived from the traditional structured NLP pipeline.
Formally, a sentence is represented as a list of tokens $X = [t_0, t_1, \cdots, t_{n-1}]$, and a labeled edge is represented as $\{s^{(1)}, s^{(2)}, \mathcal{L}\}$, where $s^{(1)} = [i^{(1)}, j^{(1)})$ and $s^{(2)} = [i^{(2)}, j^{(2)})$ (the second span is optional\footnote{For unary edge labels, such as sentence constituents, $s^{(2)}$ is omitted. See \S\ref{sec:probing_task} for a detailed description of the task.}) represent spans excluding the end word, and $\mathcal{L}$ represents a set of zero or more targets from a specific task label set $\mathcal{L}$. When inputting to the probing classifier, the first token of each span is used as the representation of that span. 
In our probing framework, the probing classifier makes predictions for unary edge labels as follows:
\begin{equation}
    \hat{Y}_{\tau}=P_{\tau}\left( \mathbf{q}_{i^{\left( 1 \right)}}^{out} \right) 
\end{equation}
For binary edge labels, the prediction of the probing classifier is expressed as follows:
\begin{equation}
    \hat{Y}_{\tau}=P_{\tau}\left( \mathbf{q}_{i^{\left( 1 \right)}}^{out}:\mathbf{q}_{i^{\left( 2 \right)}}^{out} \right) 
\end{equation}
where the input to the probing classifier is the concatenation of the first token representations of $s^{(1)}$ and $s^{(2)}$. For single-label output probing tasks, the Cross-Entropy Loss function is employed for optimization, whereas for multi-label output probing tasks, the Binary Cross-Entropy (BCE) Loss function is utilized.

\section{Probing Tasks}
\label{sec:probing_task}
The edge probing suite~\cite{tenney2018what} has been enhanced to deliver comprehensive syntactic and semantic probing tools, targeting eight core annotation tasks in natural language processing. The syntactic tasks encompass part-of-speech tagging, constituent parsing, and dependency parsing, while the semantic tasks include named entity recognition, semantic role labeling, coreference resolution, semantic proto-role labeling, and relation classification. Examples of word-level probing tasks are shown in Table~\ref{tab:word_level_example}.
Unlike traditional approaches to these tasks, the Edge Probing Suite is specifically designed to assess whether a language model's internal representations capture diverse linguistic phenomena, such as syntactic structures and semantic relationships. Its primary focus is on analyzing the model's representational capabilities, rather than solving the tasks for practical applications.



\subsection{Syntactic Tasks}
\begin{itemize}
    \item Part-of-speech tagging (POS) is a syntactic task that assigns labels, such as NOUN or VERB, to individual words in a sentence. It is classified as a unary edge task, as it involves a single span within the sentence, where the span length corresponds to one word, i.e., $s^{(1)} = [i, i + 1)$. The objective of the task is to predict the POS label for the span $s^{(1)}$.
    \item  Constituent labeling (Constituents) focuses on assigning a non-terminal label to a span in the phrase structure analysis of a sentence. For instance, it involves identifying whether a span corresponds to a noun phrase, verb phrase, or another syntactic category. Given a known constituent span $s^{(1)} = [i, j)$, the task aims to predict the appropriate label for that constituent. 
    \item Dependency labeling (Dependencies) focuses on predicting the functional relationship between two tokens. Examples include modifier-head relationships, subject-object relationships, and more. In this task, $s^{(1)} = [i, i + 1)$ represents a single token, while $s^{(2)} = [j, j + 1)$ represents the syntactic head of $s^{(1)}$. The goal is to predict the dependency relationship between token $i$ and $j$.
\end{itemize}

\subsection{Semantic Tasks}

\begin{itemize}
    \item Named entity labeling (Entities) is a task focused on identifying the category of an entity within a specified span, such as determining whether the entity represents a person, location, organization, or other types. Given an entity span $s^{(1)} = [i, j)$, the objective is to predict the corresponding entity type for that span.
    \item Semantic role labeling (SRL) analyzes sentences to identify the roles of arguments in relation to a predicate. For example, in "Mary pushed John," SRL labels "Mary" as ARG0 (initiator) and "John" as ARG1 (receiver). Given $s^{(1)}$ as the predicate and  $s^{(2)}$ as the argument, the task is to predict the role of  $s^{(1)}$, such as ARG0 or ARG1.    
    \item Coreference resolution (Coreference)  involves determining whether two mention phrases in a text refer to the same entity or event. Given $s^{(1)}$ and $s^{(2)}$ the mentions, the task is to predict whether they corefer with a binary decision.
    \item Semantic proto-role (SPR) labeling assigns fine-grained, non-exclusive semantic attributes to predicate-argument pairs, such as changes of state or awareness. For instance, in "Mary pushed John," while SRL identifies "Mary" as the pusher, SPR captures attributes like whether the pusher is aware of the action. In this task, $s^{(1)}$ represents the predicate span, $s^{(2)}$ represents the argument head, and the goal is to perform multi-label classification of the attributes characterizing the predicate-argument relationship.    
    \item Relation Classification (Rel.) aims to predict the real-world relationship between two entities from a predefined set of relationship types. In this task, $s^{(1)}$ and $s^{(2)}$ represent the given entity mentions, and the objective is to determine the type of relationship between them.
\end{itemize}

\section{Experiments}

\subsection{Experimental Setup}

\textit{1) Dataset selections:} 
We utilize annotated data from the OntoNotes 5.0 corpus\footnote{\url{https://catalog.ldc.upenn.edu/LDC2013T19}} for five of the eight tasks: part-of-speech tagging, constituent parsing, named entity recognition, semantic role labeling, and coreference resolution. We convert their original annotations into the edge probing format.
For part-of-speech tagging, the tags are extracted from the constituent parsing data in OntoNotes. For coreference resolution, as OntoNotes provides only positive examples (mentions that corefer), negative examples are created by pairing mentions that are not explicitly marked as coreferent.
Additionally, for the dependency labeling task, we choose the English Web Treebank portion of the Universal Dependencies 2.2 dataset~\cite{silveira-etal-2014-gold}. For proto-role labeling, the SPR dataset from the English Web Treebank is used \cite{rudinger-etal-2018-neural}. For relation classification, the SemEval 2010 Task 8 (SemEval) dataset is employed \cite{hendrickx-etal-2010-semeval}. The statistical information of the datasets used is provided in Table~\ref{tab:data_sta}.

\textit{2) Language models:} 
In the frozen module, we employ three LMs: BERT-base~\cite{devlin-etal-2019-bert}, T5-base~\cite{raffel2020exploring}, and Llama-3-8B~\cite{llama3modelcard}, which represent three distinct architectural frameworks: encoder-only, encoder-decoder, and decoder-only, respectively.
For the first two LMs, we directly use the output of their encoders as the initial representation of the LM, which also serves as the output of the frozen module. However, Llama-3-8B, lacking an encoder structure, is not inherently suited for text embedding tasks that require rich contextualized representations. To address this limitation, we utilize a variant of Llama-3-8B adapted through the LLM2Vec method~\cite{llm2vec}. LLM2Vec comprises three steps: 1) enabling bidirectional attention, 2) masked next token prediction, and 3) unsupervised contrastive learning. This method effectively transforms a decoder-only LM into a powerful text encoder.

\textit{3) Implementation details:}
We employ Stanza~\cite{qi-etal-2020-stanza} to parse each sentence and extract its dependency tree structure based on the Universal Dependencies (UD) framework~\cite{de-marneffe-etal-2021-universal}, providing an explicit syntactic representation.
This structure is then used as the graph input for the \textit{GNN} and \textit{Message} components.
For each probing task, we use the Adam optimizer with a batch size of 64 and train for 10 epochs. The control module is evaluated in four configurations: three based on the designed components and one that bypasses the control module, directly connecting to the classification module (the ``Without'' route in Fig.~\ref{fig:framework}).
Each component configuration employs a two-layer network for processing.
To ensure reliability, all experiments are repeated 5 times with different random seeds, and we report the average macro F1 score along with its variance across these 5 runs.


\begin{table*}[]
\centering
\caption{The overall performance comparison of control modules (GNN, Message, and MLP) against the baseline (Without) across eight probing tasks and three language models. \textbf{Bold} indicates the best-performing control module for each probing task, $\Delta$ denotes the improvement achieved by the control modules over the baseline. The MLP module, relying solely on feature-transformation operations, achieves the best results, followed by the GNN module, which incorporates both feature-transformation and message-passing mechanisms. In contrast, the Message module, relying exclusively on message-passing, exhibits mixed results, including significant declines in certain cases.}
\scalebox{0.99}{
\begin{tabular}{@{}l|rr|rr|rr@{}}
\toprule[1.2pt]
\textbf{Model}          & \multicolumn{2}{c|}{\textbf{BERT}}              & \multicolumn{2}{c|}{\textbf{T5}}                        & \multicolumn{2}{c}{\textbf{Llama-3-8B}}            \\ \midrule[1.2pt]
\textbf{Control}        & \textbf{Without}       & \textbf{GNN ($\Delta$)}                & \textbf{Without}                & \textbf{GNN ($\Delta$)}               & \textbf{Without}        & \textbf{GNN ($\Delta$)}               \\ \midrule
\textbf{Part-of-Speech} & 81.90±0.05        & 85.38±0.72 (+3.49)            & 76.82±0.07                 & 80.39±0.45 (+3.57)          & 80.75±0.43         & 83.19±0.38 (+2.43)          \\
\textbf{Constituents}   & 54.37±0.25        & 60.22±0.1 (+5.85)            & 50.28±0.11                 & 56.36±0.09 (+6.08)          & 55.63±0.06         & 58.59±0.36 (+2.96)          \\
\textbf{Dependencies}   & 57.51±0.12        & 65.37±0.27 (+7.86)            & 50.67±0.24                 & 59.52±0.07 (+8.85)          & 58.81±0.18         & 59.81±0.10 (+1.0)           \\ \midrule
\textbf{Entities}       & 85.91±0.15        & 86.7±0.38 (+0.79)           & 77.21±0.47                 & 83.55±0.88 (+6.34)          & 69.55±0.31         & 72.55±0.41 (+3.0)           \\
\textbf{SRL}            & 26.18±0.80        & 28.29±0.51 (+2.11)           & 24.35±0.04                 & 27.52±0.14 (+3.18)          & 22.23±0.12         & 23.56±0.92 (+1.33)          \\
\textbf{Coreference}    & 89.09±0.15        & 90.1±0.15 (+1.01)           & 86.66±0.09                 & 88.64±0.09 (+1.99)          & 80.16±0.15         & 82.01±0.12 (+1.85)          \\
\textbf{SPR}            & 40.35±0.11        & \textbf{42.48±0.15 (+2.12)}           & 37.73±0.37                 & \textbf{40.63±0.11 (+2.91) }& 41.86±0.47        &  42.27±0.51 (+0.41)           \\
\textbf{Rel.(SemEval)} & 67.94±0.23        & 74.2±0.28 (+6.26)         & 65.50±0.09                  & 70.19±0.11 (+4.69)          & 61.70±0.19        & \textbf{63.44±0.45 (+1.74)}  \\ \midrule
\textbf{Average}        & 62.91             & 66.59 (+3.68)                & 58.66                      & 63.35 (+4.69)                      & 58.84              & 60.68 (+1.84)                      \\ \midrule[1.2pt] \midrule
\textbf{Control}        & \textbf{Message ($\Delta$)}  & \textbf{MLP ($\Delta$)}                & \textbf{Message ($\Delta$)}          & \textbf{MLP ($\Delta$)}               & \textbf{Message ($\Delta$)}   & \textbf{MLP ($\Delta$)}               \\ \midrule
\textbf{Part-of-Speech} & 79.30±0.46 (-2.6) & \textbf{86.37±0.68 (+4.47)}  & 73.28±0.07 (-3.53)      & \textbf{81.8±0.37 (+4.99)}  & 72.87±1.19 (-7.88)  & \textbf{83.98±0.33 (+3.23)} \\
\textbf{Constituents}   & 52.44±0.07 (-1.93) & \textbf{61.23±0.07 (+6.87)}  & 48.91±0.22 (-1.37)     & \textbf{58.20±0.23 (+7.92)} & 48.56±0.10 (-7.07)  & \textbf{61.09±0.18 (+5.46)} \\
\textbf{Dependencies}   & 56.35±0.30 (-1.15) & \textbf{66.62±0.19 (+9.11)} & 51.97±0.28 (+1.3)       & \textbf{61.73±0.2 (+11.06)} & 49.62±0.07 (-9.19)  & \textbf{64.29±0.11 (+5.48)} \\ \midrule
\textbf{Entities}       & 84.44±0.20 (-1.47) & \textbf{87.29±0.12 (+1.38)}  & 82.14±0.4 (+4.93)       & \textbf{85.54±0.55 (+8.33)} & 67.14±0.06 (-2.4)   & \textbf{74.03±0.60 (+4.48)} \\
\textbf{SRL}            & 25.19±0.33 (-0.99) & \textbf{29.17±0.68 (+2.99)}  & 25.50±0.17 (+1.15)       & \textbf{28.20±0.4 (+3.85)}  & 18.81±0.32 (-3.42)  & \textbf{23.91±0.15 (+1.68)} \\
\textbf{Coreference}    & 84.08±0.04 (-5.0) & \textbf{90.84±0.13 (+1.76)}   & 82.88±0.17 (-3.78)     & \textbf{89.51±0.03 (+2.86)} & 69.85±0.27 (-10.31) & \textbf{82.82±0.07 (+2.66)} \\
\textbf{SPR}            & 41.58±0.18 (+1.23) & 42.32±0.20 (+1.97)  & 39.12±0.29 (+1.39)       &  40.52±0.08 (+2.79)       & \textbf{42.88±1.32 (+1.02)}  &  42.35±0.25 (+0.49)     \\
\textbf{Rel.(SemEval)} & 70.15±0.23 (+2.21) & \textbf{75.56±0.35 (+7.62)}  & 66.94±2.09 (+1.44)   & \textbf{73.35±2.25 (+7.84)} & 57.26±0.37 (-4.44)  & 62.76±0.36 (+1.06) \\ \midrule
\textbf{Average}        & 61.69 (-1.22)     & \textbf{67.43 (+4.52)}              & 58.84 (+0.18)                      & \textbf{64.48 (+5.82)}             & 53.37 (-5.47)              & \textbf{61.90 (+3.06)}             \\ \bottomrule[1.2pt]
\end{tabular}}
\label{tab:word-level}
\end{table*}

\begin{figure*}[t]
    \centering
    \includegraphics[width=0.74\linewidth]{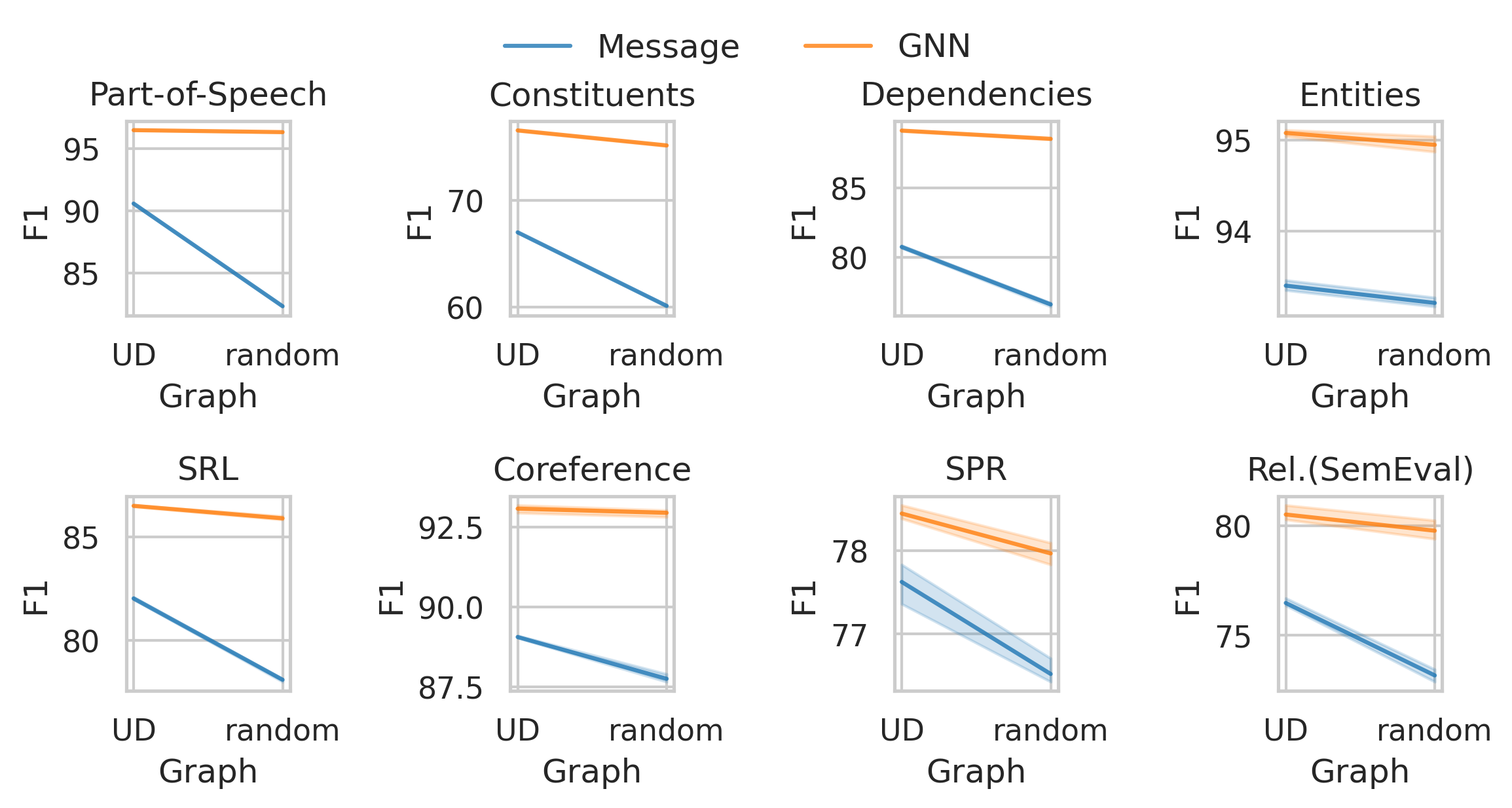}
    \caption{Performance comparison of GNN and Message control modules using standard UD graphs and random graphs. Random graphs cause a consistent performance decline, with the Message module showing greater sensitivity to structural noise than the GNN module.}
    \label{fig:random}
\end{figure*}
\begin{figure*}[t]
    \centering
    \includegraphics[width=0.85\linewidth]{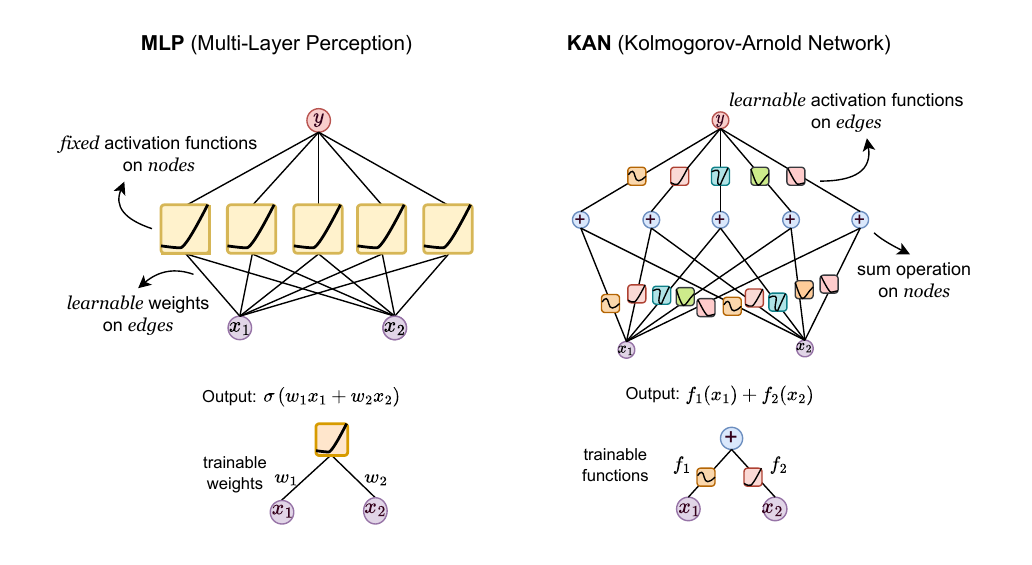}
    \caption{Overall structural differences (Top) and node-level differences (Bottom) between MLPs and KANs: MLPs use fixed activation functions on nodes, while KANs apply learnable activation functions on edges and replace linear weights with spline-based functions.}
    \label{fig:kan_model}
\end{figure*}

\subsection{Overall Results}
The overall experimental results from our probing framework on the edge probing suite are presented in Table~\ref{tab:word-level} across various tasks. The experiments involve three types of control modules—GNN, Message, and MLP—and compare their effectiveness against the baseline setup (WO, without control module). $\Delta$ indicates the improvement achieved by the control modules over the baseline.

\textit{1) General Trends:}
Across all language models and tasks, the addition of GNN or MLP control modules consistently improves performance compared to the baseline (WO). Specifically, in the BERT model, the GNN module achieves an average performance increase of 3.68, while the MLP module delivers an even larger improvement of 4.52 on average. In contrast, the Message module results in an average performance drop of 1.22. For the T5 model, all three control modules lead to performance gains, with GNN contributing an improvement of 4.69, MLP achieving the highest increase of 5.82, and Message offering a modest gain of 0.18. In the case of the Llama-3-8B model, the enhancements brought by GNN and MLP are relatively smaller, at 1.84 and 3.06, respectively, while the Message module introduces a significant negative impact, with an average performance decline of 5.37.
These results highlight the pivotal role of feature-transformation operations in GNN and MLP modules in driving performance improvements. Notably, the MLP module, despite lacking a message-passing mechanism, consistently outperforms the GNN module. This suggests that effective feature transformation alone may be sufficient to capture task-relevant information, obviating the need for the additional computational complexity introduced by explicit message-passing mechanisms.


\textit{2) Model-level Comparison:}
The variations in performance gains across different models suggest architectural differences in how these models utilize additional control mechanisms, underscoring the importance of model-module compatibility. For instance, T5 and BERT benefit more substantially from the incorporation of control modules, while Llama-3-8B exhibits comparatively smaller gains. Notably, the Message module has a pronounced negative effect on Llama-3-8B, further emphasizing the need for compatibility between control modules and model architectures.
Despite being adapted into an effective text encoder through the LLM2Vec method, decoder-only models like Llama-3-8B appear to retain representations influenced by their intrinsic autoregressive mechanisms. This residual influence makes Llama-3-8B more sensitive to the introduction of context-dependent structural information, potentially limiting its ability to fully leverage certain control modules, such as Message. These findings provide valuable insights into the interplay between model architecture and control mechanisms, offering guidance for optimizing the integration of control modules in pre-trained language models.

\textit{3) Task-specific Comparison:}
For syntactic tasks, the Message module, which depends exclusively on the message-passing mechanism, consistently shows a performance decline across all models and tasks—except for the dependency labeling task in T5. This indicates that explicit structural information can not help capture the complex syntactic relationships required by most tasks.
For semantic tasks, explicit structural information does not exhibit an entirely negative impact. Specifically, in the Semantic Proto-Role task, the use of the Message module consistently leads to performance improvements across all language models. Additionally, when leveraging this control module, performance gains are observed in four out of five semantic tasks with T5, in two tasks with BERT, and in only one task with Llama-3-8B.
These findings highlight the importance of tailoring model design to the specific demands of the task, while acknowledging the variable effects of structural information.

\section{Analysis}

\subsection{Standard UD vs Random Graph}

To further investigate the role of explicit syntactic structures in language model representations, we construct random graphs for the datasets in each probing task. These graphs are integrated into the BERT-based probing framework's GNN and Message control module, enabling a systematic comparison with the standard UD graphs.
As shown in Fig.~\ref{fig:random}, performance consistently declines across both control modules when random graphs are used, indicating that structural noise disrupts the meaningful syntactic information provided by standard UD graphs.
Notably, in nearly all probing tasks, the impact of noise introduced by random graphs is significantly greater when using the Message control module compared to the GNN control module, as reflected in a relatively larger performance drop. This suggests a difference in sensitivity to structural noise between the two control modules, with the Message control module being more sensitive.
This further supports the hypothesis that the feature-transformation operation in the GNN may mitigate the negative effects of structural noise, as it can potentially capture relevant linguistic features more effectively. This observation aligns with the overall results presented in Table~\ref{tab:word-level}.

\begin{figure*}[t]
    \centering
    \includegraphics[width=1.0\linewidth]{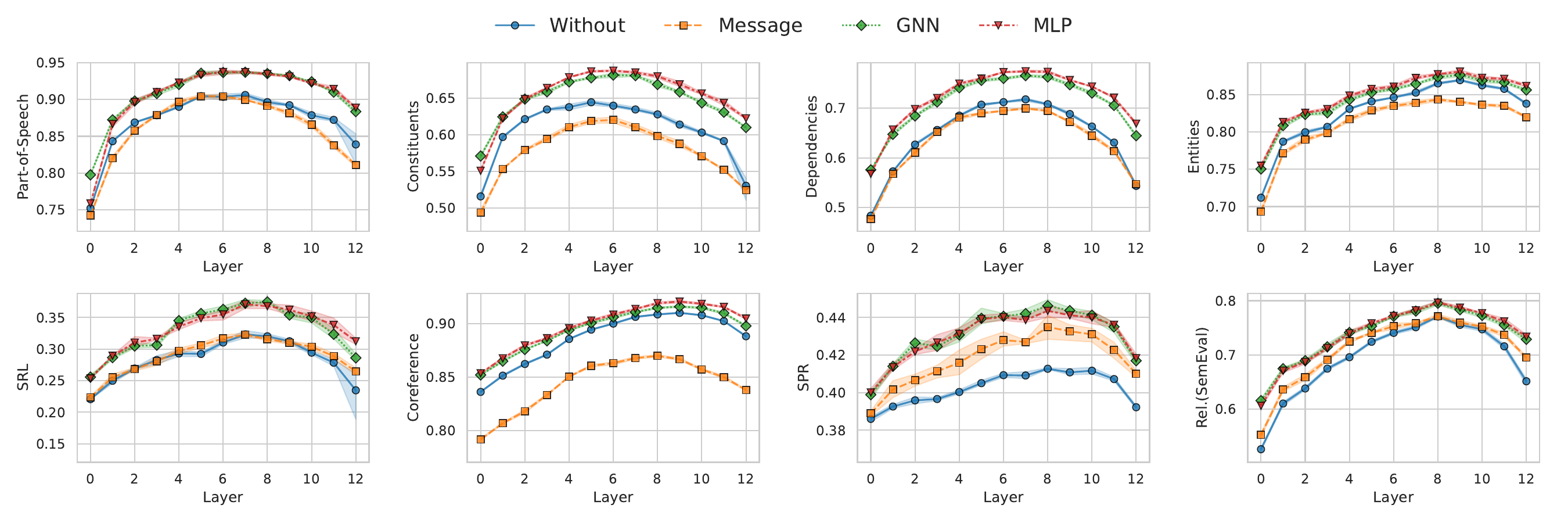}
    \caption{Probing performance across BERT hidden layers for different control modules. MLP and GNN consistently enhance performance, with MLP slightly outperforming GNN. The Message module generally performs worse, except for tasks like dependency labeling and relation extraction, highlighting the robustness of feature-transformation operations.}
    \label{fig:5layer}
\end{figure*}

\subsection{KAN as an Alternative to MLP}

\begin{table}[t]
\centering
\caption{Performance comparison of MLP and KAN across probing tasks (mean ± standard deviation, differences in parentheses). KAN performs slightly lower overall but is comparable on Entities and SPR.}
\begin{tabular}{@{}l|rr@{}}
\toprule
\textbf{Control}        & \textbf{MLP} & \textbf{KAN} \\ \midrule
\textbf{Part-of-Speech} & 86.37±0.68   & 85.96±0.66 (-0.41)  \\
\textbf{Constituents}   & 61.23±0.07   & 60.05±0.20 (-1.18)   \\
\textbf{Dependencies}   & 66.62±0.19   & 65.58±0.31 (-1.04)  \\ \midrule
\textbf{Entities}       & 87.29±0.12   & 87.12±0.21 (-0.17)  \\
\textbf{SRL}            & 29.17±0.68   & 28.29±0.29 (-0.88)  \\
\textbf{Coreference}    & 90.84±0.13   & 90.42±0.06 (-0.42)  \\
\textbf{SPR}            & 42.32±0.20   & 42.16±0.35 (-0.16)  \\
\textbf{Rel.(SemEval)}  & 75.56±0.35   & 74.73±0.13 (-0.83)   \\ \midrule
\textbf{Average}        & 67.43        & 66.79 (-0.64)       \\ \bottomrule
\end{tabular}
\label{tab:KAN_result}
\end{table}

To further validate the effectiveness of feature-transformation operations, we employ Kolmogorov-Arnold Networks (KANs)~\cite{liu2024kan} for experimental comparison. KANs have been demonstrated to be promising alternatives to MLPs. 
Figure~\ref{fig:kan_model} highlights the overall structural differences and node-level differences between these two network architectures. 
Specifically, MLPs employ fixed activation functions on neuron nodes, whereas KANs feature learnable activation functions on edges. Furthermore, KANs eliminate the use of linear weights entirely, replacing each weight parameter with a univariate function parameterized as a spline.
As shown in Table~\ref{tab:KAN_result}, when KAN is used as the control module to replace MLP, its overall performance is slightly lower than MLP but remains within an acceptable range. Notably, KAN achieves performance comparable to MLP on tasks such as Entities and SPR, highlighting its potential as a viable alternative. These findings further underscore the pivotal role of feature-transformation operations in enhancing the representation capabilities of language models. Additionally, the relatively small standard deviations observed for both KAN and MLP indicate stable performance across tasks, lending strong reliability to the reported results.
By comparing KANs and MLPs, we demonstrate the robustness of feature-transformation operations as a core mechanism for improving language model representations.

\subsection{Layer-wise Probing Analysis}
\begin{figure}[t]
    \centering
    \includegraphics[width=1\linewidth, trim=0cm 3.5cm 0cm 0cm, clip]{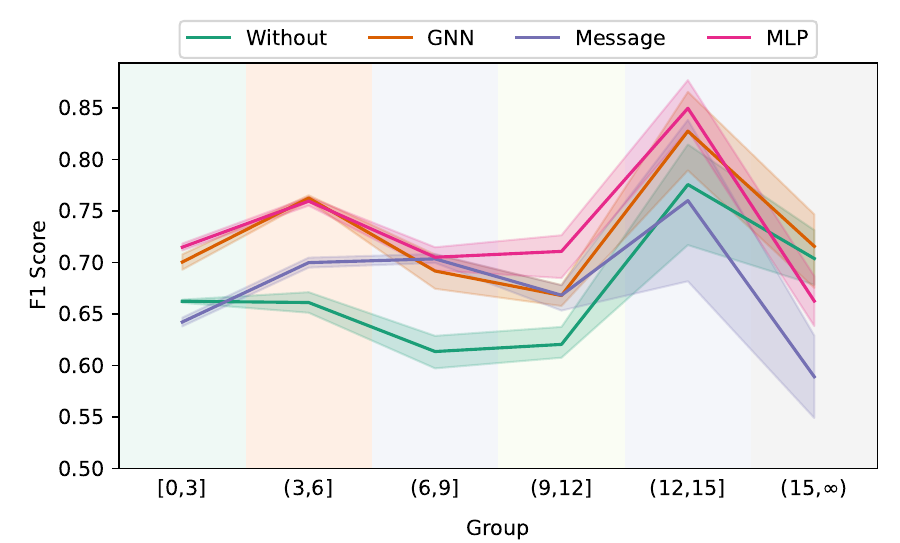}
    \caption{Probing performance across span distance groups for relation extraction. }
    \label{fig:5span}
\end{figure}

\begin{figure*}[ht]
    \centering
\includegraphics[width=1.0\linewidth]{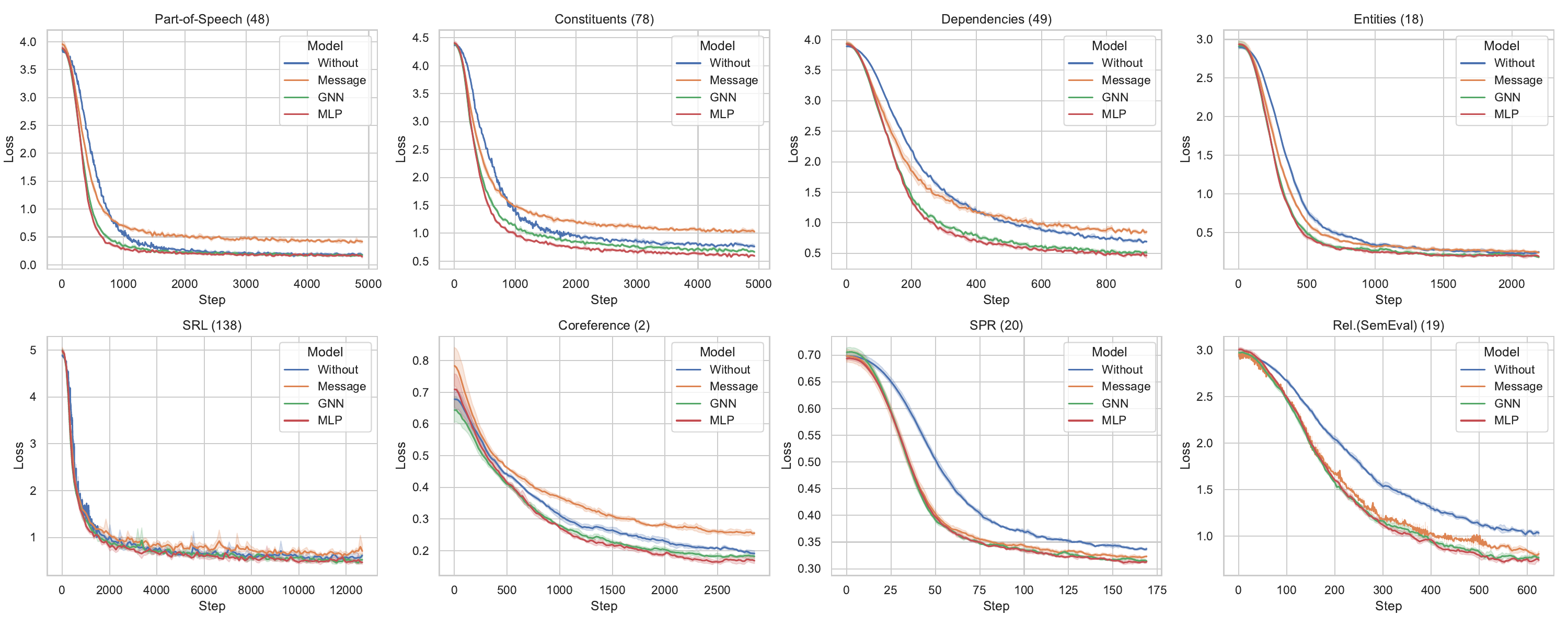}
    \caption{Loss function curves of different control modules across various probing tasks.}
    \label{fig:loss_draw_EdgeProb}
\end{figure*}
Considering the varying abilities of different layers in a language model to capture syntactic and semantic structures within a sentence, we use the hidden layer representations from each layer of BERT as inputs to the probing framework, rather than relying solely on the final layer. Similarly, we conduct experiments with different control modules on layer-wise representations to further explore the distinct roles and practical effects of feature-transformation and message-passing operations when decoupled.
The probing results are shown in Fig.~\ref{fig:5layer}.
In general, regardless of which layer's hidden representation from the language model is used as input to the probing framework, the findings align with the overall results (Table~\ref{tab:word-level}). Specifically, both the MLP and GNN modules, which incorporate feature-transformation operations, consistently enhance performance, with the MLP module slightly outperforming the GNN module. In contrast, the Message module, relying solely on message-passing operations, generally leads to performance declines, except in the dependency labeling and relation extraction probing tasks.
This demonstrates that the observed benefits of feature-transformation modules are not limited to specific representations (e.g., high-level semantic layers or low-level syntactic layers). 
These results demonstrate that the benefits of feature-transformation modules are robust and not limited to specific representations (e.g., high-level semantic layers or low-level syntactic layers). This generalizability reinforces the conclusion that feature-transformation operations are a fundamental and reliable mechanism for enhancing language model representations, independent of the input layer or representation type.


\subsection{Impact of Span Distances}
We further investigate the impact of the message-passing mechanism for explicit structural modeling and the feature-transformation operation on binary probing tasks with varying span distances. Specifically, for the relation extraction task, the test set data is grouped based on the actual distances between entity spans, and separate probing evaluation experiments are performed for each group.
As shown in Fig.~\ref{fig:5span}, the Message module does not improve performance when spans are either short ([0,3]) or long ($(12,\infty)$). This may be because short spans require little structural information, while structural information is less effective for very long spans. 
In contrast, both the GNN and MLP modules consistently enhance performance across all span distances, demonstrating that even without explicit structural modeling, MLPs effectively improve the model's ability to capture semantic information. 
Interestingly, for very large spans ($(15,\infty)$), the GNN module slightly outperforms the MLP module, suggesting that the message-passing mechanism in GNNs provides a marginal advantage in capturing long-distance relationships.


\subsection{Training Convergence Analysis}
To assess the training efficiency and performance stability of the probing framework across different control module designs, we present the loss function curves, as illustrated in Fig.~\ref{fig:loss_draw_EdgeProb}.
To evaluate the training efficiency and performance stability of the probing framework under different control module designs, we present the corresponding loss function curves, as shown in Fig.~\ref{fig:loss_draw_EdgeProb}. The results reveal that the GNN and MLP modules achieve superior performance across most tasks, with MLP outperforming GNN, as evidenced by faster convergence rates and lower final loss values. In contrast, the Without configuration exhibits the poorest performance. These findings further underscore the positive impact of feature-transformation operations in enhancing language representation capabilities, while also questioning the necessity of message-passing mechanisms that rely on explicit structural modeling. Thus, MLPs are sufficient for effectively enhancing language model representations.

\section{Conclusion}

In this paper, we challenge the prevailing assumption that explicit structural modeling through GNNs is essential for improving language model representations in NLP tasks. 
Using a modular analysis approach, we isolate and evaluate the individual contributions of key operations within GNNs, avoiding confounding effects introduced by the full architecture. 
Specifically, by decoupling the micro-architecture of GNNs and leveraging a novel probing framework, we reveal that the observed performance gains are primarily driven by feature transformation operations, rather than the message-passing mechanisms traditionally emphasized in GNN designs.
Moreover, the comparable or superior performance achieved by simpler models, such as MLPs and alternative architectures like Kolmogorov–Arnold Networks (KANs), highlights that explicit structural modeling is not always necessary. 
These findings challenge the conventional reliance on complex architectures and demonstrate the potential of simpler, more efficient models to deliver robust language representations.

\ifCLASSOPTIONcaptionsoff
  \newpage
\fi

\bibliographystyle{IEEEtran}
\bibliography{IEEEabrv,Bibliography,anthology,IEEE_reference}

\vfill

\newpage

\end{document}